\documentclass{article}
\usepackage{spconf,amsmath,graphicx}
\usepackage{booktabs}
\usepackage{multirow}
\usepackage{multicol}
\usepackage{soul}
\usepackage{color,xcolor}
\usepackage{hyperref}
\usepackage{enumitem}

\definecolor{local}{RGB}{222,64,44}
\definecolor{global}{RGB}{246,173,83}

\title{Meeting Action Item Detection with Regularized Context Modeling}
\name{Jiaqing Liu, Chong Deng, Qinglin Zhang, Qian Chen, Wen Wang}
\address{Speech Lab of DAMO Academy, Alibaba Group \\
\tt \normalsize \{mingzhai.ljq,dengchong.d,qinglin.zql,tanqing.cq,w.wang\}@alibaba-inc.com}

\usepackage[pscoord]{eso-pic}
\newcommand{\placetextbox}[3]{
\setbox0=\hbox{#3}
\AddToShipoutPictureFG*{
\put(\LenToUnit{#1\paperwidth},\LenToUnit{#2\paperheight}){\vtop{{\null}\makebox[0pt][c]{#3}}}%
}%
}%

\begin{document}

\ninept
\maketitle
\begin{abstract}
Meetings are increasingly important for collaborations. 
Action items in meeting transcripts are crucial for managing post-meeting to-do tasks, which usually are summarized laboriously. 
The Action Item Detection task aims to automatically detect meeting content associated with action items. However, datasets manually annotated with action item detection labels are scarce and in small scale.
We construct and release the first Chinese meeting corpus with manual action item annotations\footnote{\url{https://www.modelscope.cn/datasets/modelscope/Alimeeting4MUG/summary}}.
In addition, we propose a Context-Drop approach to utilize both local and global contexts by contrastive learning, and achieve better accuracy and robustness for action item detection.
We also propose a Lightweight Model Ensemble method to exploit different pre-trained models\footnote{\url{https://github.com/alibaba-damo-academy/SpokenNLP/tree/main/action-item-detection}}.  Experimental results on our Chinese meeting corpus and the English AMI corpus demonstrate the effectiveness of the proposed approaches.
\end{abstract}
\section{Introduction}
\label{sec:intro}
Due to technological advances and the pandemic, online meetings become more and more common for collaboration and information sharing. Automatic Speech Recognition (ASR) systems can convert audio recordings of meetings into transcripts. Many Natural Language Processing (NLP) tasks are conducted on meeting transcripts to automatically extract or generate important information such as summaries, decisions, and action items.  Action item refers to a task discussed in the meeting and assigned to participant(s) and expected to complete \emph{within a short time window} after the meeting~\cite{DBLP:conf/sigdial/GruensteinNP05}. The action item detection task aims to detect sentences containing information about actionable tasks in meeting transcripts. Action item detection could help users easily summarize meeting minutes, view and follow up on post-meeting to-do tasks.


Action item detection is usually modeled as a sentence-level binary classification task, to determine whether a sentence contains action items or not.
Many previous works~\cite{morgan2006automatically} explore machine learning methods and feature engineering on publicly available meeting corpora such as ICSI~\cite{janin2003icsi} and AMI~\cite{carletta2005ami}. 
Recently, with the success of the pretraining-finetuning paradigm and the revival of meeting-related research, approaches have been proposed based on pre-trained models~\cite{sachdeva2021action}, such as BERT~\cite{DBLP:conf/naacl/DevlinCLT19} and ETC~\cite{DBLP:conf/emnlp/AinslieOACFPRSW20}. 
In addition, some works~\cite{DBLP:conf/mlmi/PurverEN06} focus on detecting each element of action items independently, including task description, ownership, timeframe, and agreement.

For action item detection, existing public meeting corpora, such as the AMI meeting corpus and the ICSI meeting corpus,  are far from adequate to evaluate advanced deep learning models. We obtain 101 annotated AMI meetings with 381 action items following previous works~\cite{sachdeva2021action}. The ICSI meeting corpus comprises only 75 meetings without publicly available action item annotations. Therefore, we construct and make available a Chinese meeting corpus of 424 meetings with manual action item annotations on manual transcripts of meeting recordings (Table~\ref{tab:data}), to prompt research on action item detection.

\sethlcolor{yellow}
\begin{figure}[t!]
    \centering
    \includegraphics[width=\linewidth]{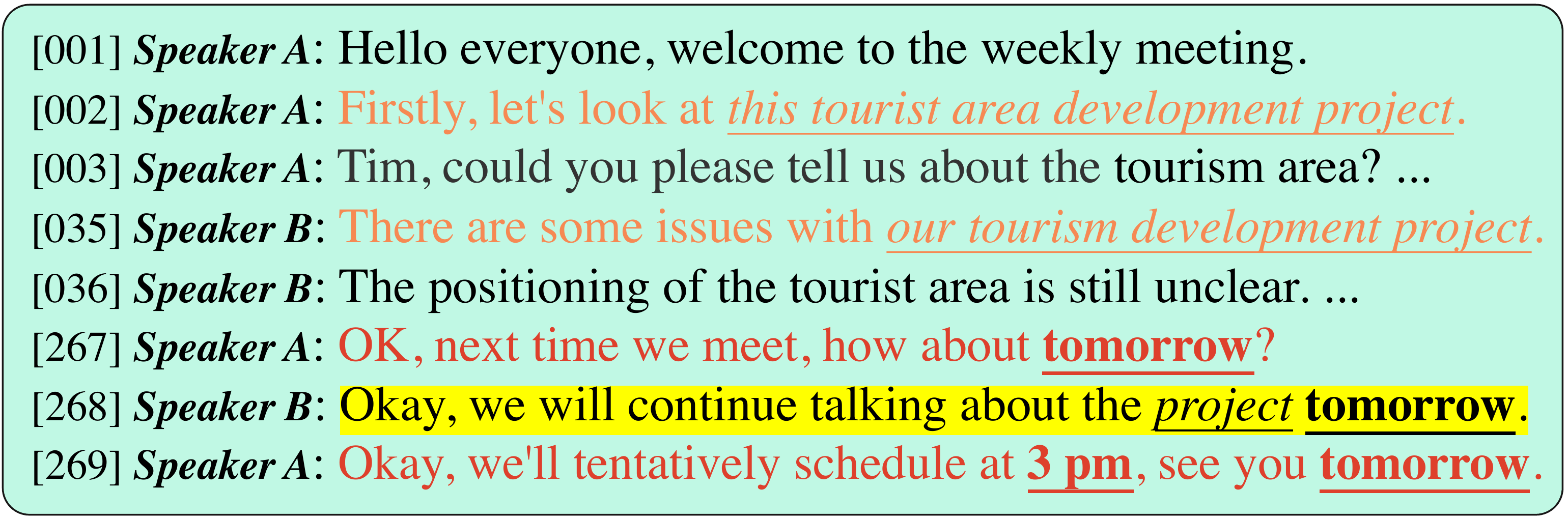}
    \caption{\small{An example of action item detection. We show the speaker and sentence id, mark the \hl{action item}, \textcolor{local}{local context} and \textcolor{global}{global context}. The \textcolor{local}{local context} provides the \underline{\textbf{timeframe}} information. And the \textcolor{global}{global context} provides the \underline{\emph{task description}} information.}
    }
    \label{fig:example}
    \vspace{-7mm}
\end{figure}



Context understanding plays a critical role in various tasks on meeting transcripts. Prior works~\cite{sachdeva2021action, mullenbach2021clip} also explore context to improve action item detection performance. 
However, most methods concatenate the focus sentence with adjacent sentences (\emph{local context}) and only achieve limited gains. As shown in Figure~\ref{fig:example}, relevant but non-contiguous sentences (\emph{global context}) also provide useful information for action items.
On the other hand, both local and global contexts may contain irrelevant information, which may distract the classifier.  We propose a novel \textbf{Context-Drop} approach to improve context modeling with regularization so that the model could focus more on the current sentence, to better exploit relevant information, and be less distracted by irrelevant information in context.

In addition, we observe that the majority voting labels are usually correct during action item annotations. Inspired by this observation, we propose a \textbf{Lightweight Model Ensemble} method to improve performance by exploiting different pre-trained models while preserving inference latency.

\placetextbox{0.5}{0.08}{\fbox{\parbox{\dimexpr\textwidth-2\fboxsep-2\fboxrule\relax}{\footnotesize \centering Accepted paper. \copyright  2023 IEEE. Personal use of this material is permitted. Permission from IEEE must be obtained for all other uses, in any current or future media, including reprinting/republishing this material for advertising or promotional purposes, creating new collective works, for resale or redistribution to servers or lists, or reuse of any copyrighted component of this work in other works.}}}

\begin{figure*}[t!]
    \centering
    \includegraphics[width=0.9\linewidth]{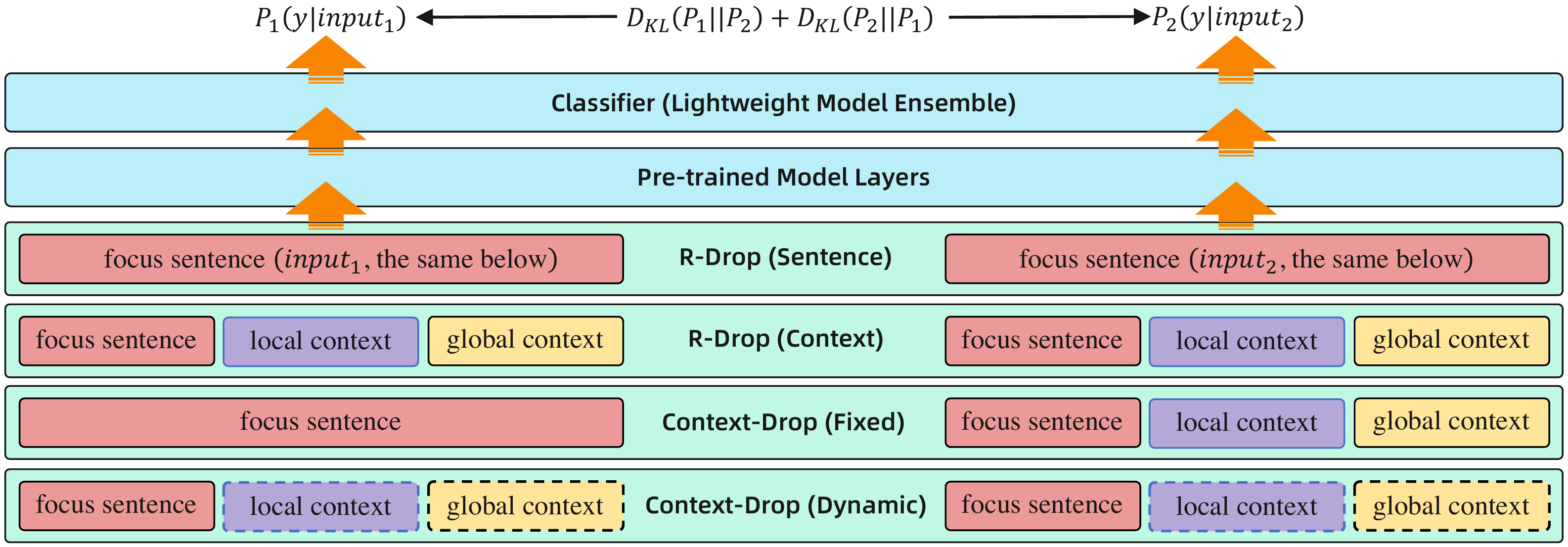}
    \caption{\small{Illustration of proposed Context-Drop  (Section~\ref{ssec:context}) and Lightweight Model Ensemble (Section~\ref{ssec:ensemble}) methods. Based on the pre-trained models, we propose the Context-Drop method to employ contextual information for action item detection. We utilize both local and global contexts to exploit as much relevant context as possible within the max sequence length of transformers. We also propose the Lightweight Model Ensemble to improve performance using different pre-trained models.}
    }
    \label{fig:model}
    \vspace{-4mm}
\end{figure*}

The contributions of our work are as follows:
\vspace{-2mm}
\begin{itemize}[leftmargin=*,noitemsep]
    \item We construct and make available a Chinese meeting corpus with action item annotations, to alleviate scarcity of resources and prompt related research. To the best of our knowledge, this is so far the largest meeting action item detection corpus.
    \item We propose a novel Context-Drop approach to improve context modeling of both local and global contexts with regularization, and achieve improvement in accuracy and robustness of action item detection for both Chinese and English meeting corpora. 
    \item We propose a Lightweight Model Ensemble approach to integrate knowledge from different pre-trained models. We achieve improvement in accuracy while preserving inference latency.
    
\end{itemize}




\section{Datasets}
\label{sec:dataset}

\vspace{-2mm}
\subsection{AMI Meeting Corpus}
\label{ssec:ami}
The AMI meeting corpus~\cite{carletta2005ami} has played an essential role in various meeting-related research. It contains 171 meeting transcripts and various types of annotations. Among 171 meetings, 145 meetings are scenario-based meetings and 26 are naturally occurring meetings. The AMI meeting corpus is a common dataset for benchmarking action item detection systems. Although there are no direct annotations for action items for this corpus, indirect annotations can be generated based on annotations of the summary. Following previous works~\cite{sachdeva2021action}, we consider dialogue acts linked to the action-related abstractive summary as positive samples for action item detection and otherwise negative samples. In this way, we obtain 101 annotated meetings with 381 action items.

\vspace{-2mm}
\subsection{Building A Large-scale Chinese Meeting Corpus}
\label{ssec:corpus}
The two common datasets for action item detection, namely the AMI meeting corpus and ICSI meeting corpus, are both far from adequate for evaluating advanced deep learning models on action item detection.
As described above, there are only 101 annotated meetings with 381 action items in the AMI meeting corpus. Another public meeting corpus, the ICSI meeting corpus, has action item annotations for 18 meetings~\cite{purver2007detecting} and is much smaller for action item detection research. Also, these annotations are no longer publicly available. 
Scarce and small-scale meeting datasets have hindered research on action item detection. To address this issue and prompt research on this topic, we construct and make available a Chinese meeting corpus, \emph{the AliMeeting-Action Corpus (denoted as AMC-A)}, with manual action item annotations on manual transcripts of meeting recordings.  We extend 224 meetings previously published in ~\cite{yu2022m2met} with additional 200 meetings. Each meeting session consists of a 15-minute to 30-minute discussion by 2-4 participants covering certain topics from a diverse set, biased towards work meetings in various industries.
All 424 meeting recordings are manually transcribed with punctuation inserted.  Semantic units ended with a manually labeled period, question mark, and exclamation are treated as \textbf{sentences} for action item annotations and modeling.

We formulate action item detection as a binary classification task and conduct sentence-level action item annotations, i.e., sentences containing action item information (task description, time frame, owner) as positive samples (labeled as 1) and otherwise negative samples (labeled as 0). As found in previous research and our experience, annotations of action items have high subjectivity and low consistency, e.g., only a Kappa coefficient of 0.36 on the ICSI corpus~\cite{purver2007detecting}.
To ease the task, we provide detailed annotation guidelines with sufficient examples.  To reduce the annotation cost, we first select candidate sentences containing both temporal expressions (e.g., ``tomorrow'') and action-related verbs (e.g., ``finish''), and highlight them in different colors.  Candidate sentences are then annotated by three annotators independently. 
During annotation, candidate sentences are presented with their context so that annotators can easily exploit context information. 
With these quality control methods, the average Kappa coefficient on AMC-A between pairs of annotators is \textbf{0.47}. For inconsistent labels from three annotators, an expert reviews the majority voting results and decides on final labels. Table~\ref{tab:data} shows that AMC-A has much more meeting sessions, total utterances, and total action items than the AMI meeting corpus and comparable avg. action items per meeting.  To the best of our knowledge, AMC-A is so far the first Chinese meeting corpus and the largest meeting corpus in any language labeled for action item detection.

\begin{table}[t!]
    \centering
     \scalebox{0.8}{
    \begin{tabular}{lrrrrr}
    \toprule
    & \multicolumn{4}{c}{\textbf{AMC-A (ours)}}  & \multirow{2}{*}{\textbf{AMI}} \\
    \cmidrule(lr){2-5}
    & \textbf{All} & \textbf{Train} & \textbf{Dev} & \textbf{Test} & \\
    
    \midrule
    \textbf{Total \# Meetings} & \textbf{424} & 295 & 65 & 64 & 101 \\
    \textbf{Total \# Utterances} & \textbf{306,846} & 213,235 & 45,869 & 47,742 & 80,298 \\
    \textbf{Total \# Action} & \textbf{1506} & 1014 & 222 & 270 & 381  \\
    \textbf{Kappa Coefficient} & 0.47 & 0.46 & 0.49 & 0.50 & /  \\
    \textbf{Avg. \# Action per Meeting} & 3.55 & 3.44 & 3.42 & 4.22 & 3.77 \\
    \textbf{Std. \# Action per Meeting} & 3.97 & 3.98 & 3.35 & 4.41 & 1.95 \\
    \bottomrule
    \end{tabular}
    }
    \caption{\small{Statistics of our Chinese AMC-A corpus and the English AMI meeting corpus studied in this work.}}
    \label{tab:data}
    \vspace{-5mm}
\end{table}

\vspace{-3mm}
\section{Method}
\label{sec:method}
\vspace{-1mm}
We formulate action item detection as a binary classification task.  Given an utterance $X$ with its context $C$, the model predicts the label $\hat{y}$, i.e., whether $X$ contains action items or not. Figure~\ref{fig:model} illustrates the two proposed approaches, \textbf{Context-Drop} (Fixed and Dynamic) and \textbf{Lightweight Model Ensemble}.  Context-Drop explores local and global contexts together with regularization. Lightweight Model Ensemble is an efficient approach for improving performance using different pre-trained models while preserving inference latency.


\subsection{Context-Drop}
\label{ssec:context}
\noindent \textbf{Local and Global Context}
Coreferences and omission of information are quite common in multi-party meetings. Relevant and supporting information may appear in adjacent sentences or non-contiguous sentences. Context understanding has played a critical role in various understanding tasks in meetings, including sentence-level action item detection. 
Relevant contexts are not limited to adjacent sentences (\emph{local context}). In real meeting scenarios, topics are usually mixed, hence discussions of a certain action item may spread in the session. We denote these relevant but non-contiguous sentences by \emph{global context} for action item detection.


Since the global context may be distant from the focus sentence, including all sentences between the global context and the focus sentence may exceed the max sequence length mandated by Transformer-based pre-trained language models (PLMs), such as BERT~\cite{DBLP:conf/naacl/DevlinCLT19} and RoBERTa~\cite{DBLP:journals/corr/abs-1907-11692}, due to their quadratic time and memory complexity to the input sequence length~\cite{DBLP:conf/nips/VaswaniSPUJGKP17}. Hence, we employ a \textit{context selection} method to retrieve the global context for each sentence. We use the cosine similarity of n-grams to measure the similarity between sentences in a document, following the n-gram overlap method~\cite{han2021modeling}.  For each sentence, we select the top-k sentences with the highest similarity scores as its global context.

\noindent \textbf{Context-Drop}
We propose a novel Context-Drop approach to improve context modeling for action item detection. 
Inspired by Contrastive Learning and R-drop~\cite{wu2021r}, Context-Drop forces the prediction probability distributions of a single sentence and the sentence with its context to be consistent with each other. We hypothesize that Context-Drop could help the model to focus more on the current sentence, to better exploit relevant information and be less distracted by irrelevant information in context, which in turn could improve the robustness and performance of the model.

We propose two types of Context-Drop, namely, \emph{Context-Drop (Fixed)} and \emph{Context-Drop (Dynamic)}. As shown in Figure~\ref{fig:model}, for Context-Drop (Fixed), $input_1$ is the focus sentence, $input_2$ is the sentence with its local/global context. For Context-Drop (Dynamic), the local/global context of the focus sentence is selected dynamically. Each sentence in context has a certain probability to be kept; otherwise, the sentence is dropped from context.
Both Context-Drop variants force the prediction probability distributions for $input_1$ (denoted $x$) and $input_2$ (denoted $x^{\prime}$) to be as close as possible, by minimizing the bidirectional Kullback-Leibler divergence as in Eqn.~\ref{equ:kl}. The overall loss is calculated as Eqn.~\ref{equ:overall}, where $\alpha$ is a hyperparameter:

\begin{align}
    \mathcal{L}^i_{\text{CE}} = & - \frac{1}{2} \log \left( P_1(y_i|x_i) \cdot P_2(y_i|x^{\prime}_i) \right) \\
    \begin{split}
        \mathcal{L}^i_{\text{KL}} = & \frac{1}{2} \left( \mathcal{D}_{\text{KL}} \left( P_1(y_i|x_i) || P_2(y_i|x^{\prime}_i) \right) \right. \\ 
        & + \left. \mathcal{D}_{\text{KL}} \left( P_2(y_i|x^{\prime}_i) || P_1(y_i|x_i) \right) \right) \label{equ:kl}
    \end{split} \\
    \mathcal{L}^i = & \mathcal{L}^i_{\text{CE}}  + \alpha \cdot \mathcal{L}^i_{\text{KL}} \label{equ:overall}
\end{align}

Context-Drop (Dynamic) makes the contrast between samples more flexible.   When all contexts are dropped for both $input_1$ and $input_2$, Context-Drop (Dynamic) works equivalently to the \emph{R-Drop (Sentence)} method in Figure~\ref{fig:model}.  When all contexts are kept for both $input_1$ and $input_2$, the approach works equivalently to \emph{R-Drop (Context)}. When all contexts are dropped for $input_1$ and all contexts are kept for $input_2$, the approach works equivalently to Context-Drop (Fixed). Hence, Context-Drop (Dynamic) could be considered as a generalization of the other three methods in Figure~\ref{fig:model}.

\begin{table}[t!]
    \centering
    \begin{tabular}{llc}
    \toprule
    \textbf{Model} & \textbf{Modeling Task} & \textbf{AMC-A F$_1$} \\
    \midrule
    BERT & sentence classification & 64.76\footnotesize{$\pm$0.98} \\
    Longformer & sequence labeling & 65.35\footnotesize{$\pm$1.33} \\
    StructBERT & sentence classification & \textbf{67.84}\footnotesize{$\pm$1.20} \\
    \bottomrule
    \end{tabular}
    \caption{\small{Positive F$_1$ on the \textbf{Test} set of our AMC-A corpus using different pre-trained language models with different modeling tasks.}}
    \label{tab:baseline}
\end{table}

\begin{table}[t!]
    \centering
    \begin{tabular}{lcc}
    \toprule
    \textbf{Input Method} & \textbf{AMC-A F$_1$} & \textbf{AMI F$_1$} \\
    
    \midrule
    sentence & 67.84\footnotesize{$\pm$1.20} & 38.67\footnotesize{$\pm$1.25} \\
    w/\ R-Drop & 68.77\footnotesize{$\pm$0.82} & 39.26\footnotesize{$\pm$1.70} \\
    
    \midrule
    + local context & 68.50\footnotesize{$\pm$1.21} & 41.03\footnotesize{$\pm$1.42} \\
    w/\ R-Drop & 68.79\footnotesize{$\pm$0.42} & \underline{42.72}\footnotesize{$\pm$0.74} \\
    w/\ $\text{Context-Drop}_{\text{fixed}}$ & 69.15\footnotesize{$\pm$0.91} & \textbf{43.12}\footnotesize{$\pm$0.74} \\
    \quad\;w/o KL loss & 68.23\footnotesize{$\pm$1.11} & 40.71\footnotesize{$\pm$1.78} \\
    w/\ $\text{Context-Drop}_{\text{dynamic}}$ & 69.53\footnotesize{$\pm$0.75} & 42.05\footnotesize{$\pm$0.31} \\
    \quad\;w/o KL loss  & 67.97\footnotesize{$\pm$0.53} & 41.44\footnotesize{$\pm$2.29} \\
    
    \midrule
    + global context & 67.99\footnotesize{$\pm$1.86} & 35.82\footnotesize{$\pm$1.11} \\
    w/\ R-Drop & 69.80\footnotesize{$\pm$1.14} & 37.88\footnotesize{$\pm$1.04} \\
    w/\ $\text{Context-Drop}_{\text{fixed}}$ & 69.07\footnotesize{$\pm$0.57} & 39.23\footnotesize{$\pm$0.73} \\
    w/\ $\text{Context-Drop}_{\text{dynamic}}$ & \underline{70.48}\footnotesize{$\pm$0.63} & 41.25\footnotesize{$\pm$1.76} \\
    
    \midrule
    + local \& global context & 69.09\footnotesize{$\pm$1.23} & 41.31\footnotesize{$\pm$1.51} \\
    w/\ R-Drop & 68.72\footnotesize{$\pm$1.04} & 40.75\footnotesize{$\pm$1.28} \\
    w/\ $\text{Context-Drop}_{\text{fixed}}$ & 69.28\footnotesize{$\pm$0.95} & 38.66\footnotesize{$\pm$0.77} \\
    w/\ $\text{Context-Drop}_{\text{dynamic}}$ & \textbf{70.82}\footnotesize{$\pm$1.33} & 41.50\footnotesize{$\pm$1.52} \\
    
    \bottomrule
    \end{tabular}
    \caption{\small{Positive F$_1$ on the \textbf{Test} sets of our AMC-A corpus and the AMI meeting corpus. All experiments fine-tune the pre-trained Chinese and English StructBERT models respectively. We compare the performance of different input methods (the single focus sentence or the focus sentence with its local/global context) and different training loss, including the standard CE loss by default, with R-Drop, and with the two variations of Context-Drop (Section~\ref{ssec:context}).}}
    \label{tab:metric}
\end{table}

\begin{table}[t!]
    \centering
    \begin{tabular}{llc}
    \toprule
    \textbf{Model Layers} & \textbf{Pooler Layer} & \textbf{AMC-A F$_1$} \\
    
    \midrule
    \multirow{2}{*}{StructBERT} & StructBERT & 67.84\footnotesize{$\pm$1.20} \\
    & RoBERTa & 68.36\footnotesize{$\pm$0.93} \\
    
    \midrule
    \multirow{2}{*}{RoBERTa} & RoBERTa & 66.87\footnotesize{$\pm$0.44} \\
    & StructBERT & 67.25\footnotesize{$\pm$0.93} \\
    
    \bottomrule
    \end{tabular}
    \caption{\small{Positive F$_1$ on the \textbf{Test} set of our AMC-A corpus, from fine-tuning pre-trained StructBERT and RoBERTa models and the hybrid model using Lightweight Model Ensemble (Section~\ref{ssec:ensemble}).}}
    \label{tab:ensemble}
    \vspace{-6mm}
\end{table}

    
    
    

\subsection{Lightweight Model Ensemble}
\label{ssec:ensemble}
During action item annotation, we observe that for inconsistent labels from three annotators, the majority voting results are usually correct despite the relatively low inter-annotator agreement, as the expert only modifies 5\%-10\% of the majority voting labels. Inspired by this observation, we explore model ensemble, a common approach for improving performance. In this work, we propose a Lightweight Model Ensemble approach, which improves accuracy while preserving inference latency.  Conventionally, we initialize each layer of a classification model with parameters from the same pre-trained model. In our Lightweight Model Ensemble approach, we initialize encoder layers of the action item detection model $\theta_C$ using parameters from one pre-trained model $\theta_A$ and initialize the pooler layer of $\theta_C$ using the pooler layer parameters from another pre-trained model $\theta_B$. We then fine-tune $\theta_C$ with the cross-entropy loss on the meeting corpus. In this way, we integrate knowledge from different pre-trained models efficiently, without increasing the overall number of parameters and slowing down inference.



\vspace{-2mm}
\section{Experiments}
\label{sec:experiments}
\vspace{-2mm}
\subsection{Datasets and Metrics}
\label{ssec:dataset}
\vspace{-2mm}
We use both the AMI meeting corpus and our AMC-A corpus. 
We partition the AMI meeting corpus following the official scenario-only dataset partitioning\footnote{\url{https://groups.inf.ed.ac.uk/ami/corpus/datasets.shtml}}.
We partition AMC-A into train/dev/test sets with a ratio of 70:15:15, respectively. Considering the sparsity of positive samples, we report positive F$_1$ as the evaluation metric.


\vspace{-4mm}
\subsection{Baseline and Implementation Details}
\label{ssec:implement}
\vspace{-2mm}

To evaluate our proposed methods on the AMC-A and AMI datasets, we use the following strong baseline pre-trained models, namely, BERT~\cite{DBLP:conf/naacl/DevlinCLT19}\footnote{\url{https://github.com/google-research/bert}}, RoBERTa~\cite{DBLP:journals/corr/abs-1907-11692}, StructBERT~\cite{wang2019structbert}\footnote{\url{https://modelscope.cn/models/damo/nlp\_structbert\_backbone\_base\_std}}, and Longformer~\cite{beltagy2020longformer} which provides efficient long-sequence modeling.  For RoBERTa, We use the pre-trained Chinese RoBERTa-wwm-ext model\cite{cui-etal-2020-revisiting}\footnote{\url{https://github.com/ymcui/Chinese-BERT-wwm}}.  For Longformer, we use the pre-trained Erlangshen-Longformer-110M\cite{fengshenbang}\footnote{\url{https://github.com/IDEA-CCNL/Fengshenbang-LM}} to model action item detection as a sequence labeling task and use a fixed sliding window with size 4096 and allow one sentence overlap. The sentence labeling task takes multiple sentences as input and outputs the probabilities for every sentence.
For BERT, StructBERT, and RoBERTa, we model action item detection as a sentence classification task and truncate input to 128 tokens. The sentence classification task takes a sentence as input and outputs the probabilities for the sentence.
We compare our Context-Drop approach to R-Drop~\cite{wu2021r}. R-Drop forces the predicted probability distribution of the same sample after two dropouts to be as close as possible.
We compare the performance of  R-Drop with sentence-level inputs and context-level inputs (Figure~\ref{fig:model}).

We use TensorFlow and PyTorch to implement all models.
All PLMs used are of BERT base size.
The batch size is 32 and the dropout rate is 0.3. For each experiment in this paper, we run 5 times with different random seeds; for each run, we conduct a grid search among $\{1e-5,2e-5\}$ learning rate and $\{2, 3\}$ epochs on the dev set. We then report the mean and standard deviation of the best results from 5 runs. The weight $\alpha$ of KL divergence loss is set to 4.0 for R-Drop and 1.0 for Context-Drop by optimizing positive F$_1$ on the dev set.
For each sentence, we use its preceding sentence and following sentence as \emph{local context}, and select the top-2 most similar sentences to this sentence as \emph{global context} (see Section~\ref{ssec:context} for details).
The probability to keep contextual sentences is 50\% for \textit{local} or \textit{global contexts}, and 70\% for \textit{local \& global contexts}.
Following setups in prior works, no sampling methods are applied.

\vspace{-4mm}
\subsection{Results and Analysis}
\label{ssec:result}
\vspace{-2mm}
As shown in Table~\ref{tab:baseline}, we compare different PLMs with different modeling tasks. When modeling action item detection as a sentence classification task, StructBERT outperforms BERT with a remarkable gain of \textbf{+3.08} on positive F$_1$. The word structural pre-training objective of StructBERT reconstructs tokens in the correct order from the shuffled trigrams. This could improve its robustness to disordered sentences, which is quite common in spoken languages, and in turn improve its performance of meeting action item detection. We formulate action item detection as a sequence labeling task to exploit the advantage of Longformer in long-sequence modeling. However, we only observe limited improvement from Longformer over BERT, 0.59 gain on positive F$_1$. Therefore, we formulate action item detection as a sentence classification task and use StructBERT as the pre-trained model for evaluating Context-Drop.

As shown in Table~\ref{tab:metric}, based on the baseline StructBERT, we compare various contrastive learning methods using different contexts (Figure~\ref{fig:model}). On AMC-A, when not using contrastive learning methods, i.e., no w/ R-Drop nor w/ Context-Drop, the baseline using both local and global context performs the best (69.09), followed by the baseline using the local context (68.50). We observe the same trend on AMI.
This indicates that adjacent contextual sentences do provide useful information. Global context provides complementary information and a combination of global and local context achieves further improvement. 
On AMC-A, when using different contrastive learning methods, the configuration of using the focus sentence and local \& global context as input with $\text{Context-Drop}_{\text{dynamic}}$ achieves the best performance (70.82), outperforming the baseline using the sentence as input without contrastive learning (67.84) by \textbf{+2.98} absolute on positive F$_1$, and also outperforming R-Drop (68.72) by \textbf{+2.10} absolute gain. 
On AMI, sentence+local context w/ $\text{Context-Drop}_{\text{fixed}}$ (43.12) also outperforms the baseline sentence input (38.67) and R-Drop (42.72).
These results confirm our hypothesis that Context-Drop could help the model to focus more on the current sentence, exploit relevant information in context and be less distracted by irrelevant information.  Moreover, a reduction in the standard deviations shows that Context-Drop improves the stability and robustness of the model. For different contexts, $\text{Context-Drop}_{\text{dynamic}}$ outperforms $\text{Context-Drop}_{\text{fixed}}$ in most cases, which suggests that the flexible and dynamic contrastive learning method can achieve better performance.

We also conduct ablation analysis on Context-Drop, as in the second group of Table~\ref{tab:metric}.  Without the regularization loss of KL divergence (denoted KL loss), Context-Drop can be regarded as a data augmentation method using fixed or dynamically selected context. On AMC-A and AMI, for both $\text{Context-Drop}_{\text{fixed}}$ and  $\text{Context-Drop}_{\text{dynamic}}$, w/o KL loss degrades the performance, which indicates contrastive learning is important for gains. 
With the regularization loss, the model could better focus on the current sentence and be less distracted by irrelevant information in context.

As shown in Table~\ref{tab:ensemble}, we compare the performance of applying  Lightweight Model Ensemble integrating various pre-trained models using the sentence input. StructBERT encoder with RoBERTa pooler layer parameters achieves \textbf{+0.52} absolute gain and RoBERTa encoder with StructBERT pooler layer parameters achieves \textbf{+0.38} absolute gain. These results show that Lightweight Model Ensemble could integrate knowledge from different models and achieve better performance without increasing the overall number of parameters.

\vspace{-3mm}
\section{Conclusion}
\label{sec:conclusion}
\vspace{-2mm}
We construct and make available the first Chinese meeting corpus with action item annotations, to alleviate the scarcity of resources and prompt research on meeting action item detection. We propose Context-Drop to exploit both local and global contexts with regularization. On both our meeting corpus and English AMI meeting corpus, Context-Drop improves the accuracy and robustness of action item detection. We also propose Lightweight Model Ensemble and achieve improvement. In future work, we plan to refine Lightweight Model Ensemble and investigate its efficacy on other tasks as well as combining Context-Drop and Lightweight Model Ensemble.

\vfill\pagebreak

\bibliographystyle{IEEEbib}
\bibliography{main}

\begin{thebibliography}{10}

\bibitem{DBLP:conf/sigdial/GruensteinNP05}
Alexander Gruenstein, John Niekrasz, and Matthew Purver,
\newblock ``Meeting structure annotation: Data and tools,''
\newblock in {\em Proceedings of the 6th SIGdial Workshop on Discourse and
  Dialogue, SIGdial 2005, Lisbon, Portugal, 2-3 September 2005}, Laila
  Dybkj{\ae}r and Wolfgang Minker, Eds. 2005, pp. 117--127, Special Interest
  Group on Discourse and Dialogue (SIGdial).

\bibitem{morgan2006automatically}
William Morgan, Pi-Chuan Chang, Surabhi Gupta, and Jason Brenier,
\newblock ``Automatically detecting action items in audio meeting recordings,''
\newblock in {\em Proceedings of the 7th SIGdial Workshop on Discourse and
  Dialogue}, 2006, pp. 96--103.

\bibitem{janin2003icsi}
Adam Janin, Don Baron, Jane Edwards, Dan Ellis, David Gelbart, Nelson Morgan,
  Barbara Peskin, Thilo Pfau, Elizabeth Shriberg, Andreas Stolcke, et~al.,
\newblock ``The icsi meeting corpus,''
\newblock in {\em 2003 IEEE International Conference on Acoustics, Speech, and
  Signal Processing, 2003. Proceedings.(ICASSP'03).} IEEE, 2003, vol.~1, pp.
  I--I.

\bibitem{carletta2005ami}
Jean Carletta, Simone Ashby, Sebastien Bourban, Mike Flynn, Mael Guillemot,
  Thomas Hain, Jaroslav Kadlec, Vasilis Karaiskos, Wessel Kraaij, Melissa
  Kronenthal, et~al.,
\newblock ``The ami meeting corpus: A pre-announcement,''
\newblock in {\em International workshop on machine learning for multimodal
  interaction}. Springer, 2005, pp. 28--39.

\bibitem{sachdeva2021action}
Kishan Sachdeva, Joshua Maynez, and Olivier Siohan,
\newblock ``Action item detection in meetings using pretrained transformers,''
\newblock in {\em 2021 IEEE Automatic Speech Recognition and Understanding
  Workshop (ASRU)}. IEEE, 2021, pp. 861--868.

\bibitem{DBLP:conf/naacl/DevlinCLT19}
Jacob Devlin, Ming{-}Wei Chang, Kenton Lee, and Kristina Toutanova,
\newblock ``{BERT:} pre-training of deep bidirectional transformers for
  language understanding,''
\newblock in {\em Proceedings of the 2019 Conference of the North American
  Chapter of the Association for Computational Linguistics: Human Language
  Technologies, {NAACL-HLT} 2019, Minneapolis, MN, USA, June 2-7, 2019, Volume
  1 (Long and Short Papers)}, Jill Burstein, Christy Doran, and Thamar Solorio,
  Eds. 2019, pp. 4171--4186, Association for Computational Linguistics.

\bibitem{DBLP:conf/emnlp/AinslieOACFPRSW20}
Joshua Ainslie, Santiago Onta{\~{n}}{\'{o}}n, Chris Alberti, Vaclav Cvicek,
  Zachary Fisher, Philip Pham, Anirudh Ravula, Sumit Sanghai, Qifan Wang, and
  Li~Yang,
\newblock ``{ETC:} encoding long and structured inputs in transformers,''
\newblock in {\em Proceedings of the 2020 Conference on Empirical Methods in
  Natural Language Processing, {EMNLP} 2020, Online, November 16-20, 2020},
  Bonnie Webber, Trevor Cohn, Yulan He, and Yang Liu, Eds. 2020, pp. 268--284,
  Association for Computational Linguistics.

\bibitem{DBLP:conf/mlmi/PurverEN06}
Matthew Purver, Patrick Ehlen, and John Niekrasz,
\newblock ``Detecting action items in multi-party meetings: Annotation and
  initial experiments,''
\newblock in {\em Machine Learning for Multimodal Interaction, Third
  International Workshop, {MLMI} 2006, Bethesda, MD, USA, May 1-4, 2006,
  Revised Selected Papers}, Steve Renals, Samy Bengio, and Jonathan~G. Fiscus,
  Eds. 2006, vol. 4299 of {\em Lecture Notes in Computer Science}, pp.
  200--211, Springer.

\bibitem{mullenbach2021clip}
James Mullenbach, Yada Pruksachatkun, Sean Adler, Jennifer Seale, Jordan
  Swartz, Greg McKelvey, Hui Dai, Yi~Yang, and David Sontag,
\newblock ``Clip: A dataset for extracting action items for physicians from
  hospital discharge notes,''
\newblock in {\em Proceedings of the 59th Annual Meeting of the Association for
  Computational Linguistics and the 11th International Joint Conference on
  Natural Language Processing (Volume 1: Long Papers)}, 2021, pp. 1365--1378.

\bibitem{purver2007detecting}
Matthew Purver, John Dowding, John Niekrasz, Patrick Ehlen, Sharareh
  Noorbaloochi, and Stanley Peters,
\newblock ``Detecting and summarizing action items in multi-party dialogue,''
\newblock in {\em Proceedings of the 8th SIGdial Workshop on Discourse and
  Dialogue}, 2007, pp. 18--25.

\bibitem{yu2022m2met}
Fan Yu, Shiliang Zhang, Yihui Fu, Lei Xie, Siqi Zheng, Zhihao Du, Weilong
  Huang, Pengcheng Guo, Zhijie Yan, Bin Ma, et~al.,
\newblock ``M2met: The icassp 2022 multi-channel multi-party meeting
  transcription challenge,''
\newblock in {\em ICASSP 2022-2022 IEEE International Conference on Acoustics,
  Speech and Signal Processing (ICASSP)}. IEEE, 2022, pp. 6167--6171.

\bibitem{DBLP:journals/corr/abs-1907-11692}
Yinhan Liu, Myle Ott, Naman Goyal, Jingfei Du, Mandar Joshi, Danqi Chen, Omer
  Levy, Mike Lewis, Luke Zettlemoyer, and Veselin Stoyanov,
\newblock ``Roberta: {A} robustly optimized {BERT} pretraining approach,''
\newblock {\em CoRR}, vol. abs/1907.11692, 2019.

\bibitem{DBLP:conf/nips/VaswaniSPUJGKP17}
Ashish Vaswani, Noam Shazeer, Niki Parmar, Jakob Uszkoreit, Llion Jones,
  Aidan~N. Gomez, Lukasz Kaiser, and Illia Polosukhin,
\newblock ``Attention is all you need,''
\newblock in {\em Advances in Neural Information Processing Systems 30: Annual
  Conference on Neural Information Processing Systems 2017, December 4-9, 2017,
  Long Beach, CA, {USA}}, Isabelle Guyon, Ulrike von Luxburg, Samy Bengio,
  Hanna~M. Wallach, Rob Fergus, S.~V.~N. Vishwanathan, and Roman Garnett, Eds.,
  2017, pp. 5998--6008.

\bibitem{han2021modeling}
Rujun Han, Luca Soldaini, and Alessandro Moschitti,
\newblock ``Modeling context in answer sentence selection systems on a latency
  budget,''
\newblock in {\em Proceedings of the 16th Conference of the European Chapter of
  the Association for Computational Linguistics: Main Volume}, 2021, pp.
  3005--3010.

\bibitem{wu2021r}
Lijun Wu, Juntao Li, Yue Wang, Qi~Meng, Tao Qin, Wei Chen, Min Zhang, Tie-Yan
  Liu, et~al.,
\newblock ``R-drop: Regularized dropout for neural networks,''
\newblock {\em Advances in Neural Information Processing Systems}, vol. 34, pp.
  10890--10905, 2021.

\bibitem{wang2019structbert}
Wei Wang, Bin Bi, Ming Yan, Chen Wu, Jiangnan Xia, Zuyi Bao, Liwei Peng, and
  Luo Si,
\newblock ``Structbert: Incorporating language structures into pre-training for
  deep language understanding,''
\newblock in {\em International Conference on Learning Representations}, 2019.

\bibitem{beltagy2020longformer}
Iz~Beltagy, Matthew~E Peters, and Arman Cohan,
\newblock ``Longformer: The long-document transformer,''
\newblock {\em arXiv preprint arXiv:2004.05150}, 2020.

\bibitem{cui-etal-2020-revisiting}
Yiming Cui, Wanxiang Che, Ting Liu, Bing Qin, Shijin Wang, and Guoping Hu,
\newblock ``Revisiting pre-trained models for {C}hinese natural language
  processing,''
\newblock in {\em Proceedings of the 2020 Conference on Empirical Methods in
  Natural Language Processing: Findings}, Online, Nov. 2020, pp. 657--668,
  Association for Computational Linguistics.

\bibitem{fengshenbang}
Junjie Wang, Yuxiang Zhang, Lin Zhang, Ping Yang, Xinyu Gao, Ziwei Wu, Xiaoqun
  Dong, Junqing He, Jianheng Zhuo, Qi~Yang, Yongfeng Huang, Xiayu Li, Yanghan
  Wu, Junyu Lu, Xinyu Zhu, Weifeng Chen, Ting Han, Kunhao Pan, Rui Wang, Hao
  Wang, Xiaojun Wu, Zhongshen Zeng, Chongpei Chen, Ruyi Gan, and Jiaxing Zhang,
\newblock ``Fengshenbang 1.0: Being the foundation of chinese cognitive
  intelligence,''
\newblock {\em CoRR}, vol. abs/2209.02970, 2022.

\end{thebibliography}

\end{document}